\documentclass[letterpaper]{article}
\usepackage{flairs}
\usepackage{times}
\usepackage{helvet}
\usepackage{courier}
\usepackage{graphicx}
\usepackage{amsmath}
\usepackage{algorithm}
\usepackage{algpseudocode}
\usepackage{xcolor}
\usepackage{float}
\usepackage{amsfonts}
\usepackage{paralist}
\begin{document}
\title{Smart Sampling: Self-Attention and Bootstrapping\\ for Improved Ensembled Q-Learning}

 \author{
Muhammad Junaid Khan\\
Dept. of Computer Science\\
University of Central Florida\\
Orlando, USA\\
muhammad.junaid.khan@ucf.edu\\
\And 
Syed Hammad Ahmed\\
Dept. of Computer Science\\
University of Central Florida\\
Orlando, USA\\
syed.hammad.ahmed@ucf.edu
\And 
Gita Sukthankar\\
Dept. of Computer Science\\
University of Central Florida\\
Orlando, USA\\
gita.sukthankar@ucf.edu
}
\maketitle
\begin{abstract}
\begin{quote}
We present a novel method aimed at enhancing the sample efficiency of ensemble Q learning. Our proposed approach integrates multi-head self-attention into the ensembled Q networks while bootstrapping the state-action pairs ingested by the ensemble. This not only results in performance improvements over the original REDQ \cite{redq} and its variant DroQ \cite{droq}, thereby enhancing Q predictions, but also effectively reduces both the average normalized bias and standard deviation of normalized bias within Q-function ensembles. Importantly, our method also performs well even in scenarios with a low update-to-data (UTD) ratio. Notably, the implementation of our proposed method is straightforward, requiring minimal modifications to the base model.  
\end{quote}
\end{abstract}

\section{Introduction}

Designing reinforcement learning algorithms that learn rapidly with limited data remains a significant research challenge.  While previous approaches have solved complex control tasks, they require huge amount of samples for training and solving the task \cite{openai,meta-policy}. Newer techniques focus on achieving a high update-to-data (UTD) ratio which ensures fewer environmental interactions but high sample efficiency. RL methods such as Model-based Policy Optimization (MBPO) \cite{mbpo} achieve a high UTD ratio by using both real data and fake data generated by the model to achieve sample efficiency. In contrast, the model-free Soft Actor Critic (SAC) \cite{sac} approach has a relatively low UTD ratio of 1.

Two recent model-free ensemble methods use a very high UTD ratio to achieve greater sample efficiency: REDQ \cite{redq} and DroQ \cite{droq}.  Having a high UTD ratio risks creating estimation bias since many model updates are made after a small number of environmental interactions. To prevent this, REDQ employs a Q-function ensemble and reduces the estimation bias by employing in-target minimization over a subset of the ensemble. Similarly, DroQ, which is based on REDQ, injects model uncertainty into a smaller ensemble by using Q-learners with dropout and layer normalization to minimize estimation bias.

Our proposed method uses bootstrapping~\cite{bootstrap} to generate the samples used by the Q-learning ensemble.  Several recent deep RL algorithms have leveraged bootstrapping to improve sample complexity but not within ensemble learners. \citeauthor{bootstrap_states}  (\citeyear{bootstrap_states}) bootstrap experience trajectory clusters to improve agent generalization. \citeauthor{off_policy_bs} (\citeyear{off_policy_bs}) tackle the problem of off-policy learning with covariate or distribution shift by using bootstrapping to reweight and resample the data from the behavior policy in order to align it with target policy while reducing the bias introduced due to importance sampling.  

Here we use multi-head self attention within the individual Q-learners. Self-attention was first introduced by \citeauthor{transformer} (\citeyear{transformer}) and has been effectively employed in all transformer-based approaches in  natural language tasks. Recently, many deep RL approaches have showed its effectiveness in both single-agent and multi-agent problems \cite{decisionTransformer,updet,transformerforgames,transmix}.  

This paper introduces a variant of the REDQ and DroQ methods which, in addition to an ensemble of Q-functions, dropout layer, and layer normalization, exploits multi-head self-attention (referred as \emph{MHA}), identity connections, and bootstrapping mechanisms to further improve performance and reduce estimation bias. Our approach provides the following benefits:
\begin{compactitem}
    \item With bootstrapping we create multiple sub-samples of the agent's experiences, thus helping the agent to utilize its experiences effectively which leads to better state space exploration. 
    \item In addition, MHA effectively captures the temporal relationship between sub-sampled state-action pairs while taking into account the future states and rewards, and learning a variety of dependencies amongst state-action pairs. 
    \item Experiments show that our method effectively improves sample efficiency while reducing the estimation error.  We achieve comparable performance to REDQ even with low UTD settings. 
\end{compactitem}
We demonstrate the performance of our algorithm vs.\ REDQ and DroQ in four challenging OpenAI Gym environments: Ant-v2, Hopper-v2, Humanoid-v2, and Walker2d-v2.

\begin{figure*}[ht]
    \centering
    \includegraphics[width=0.92\textwidth]{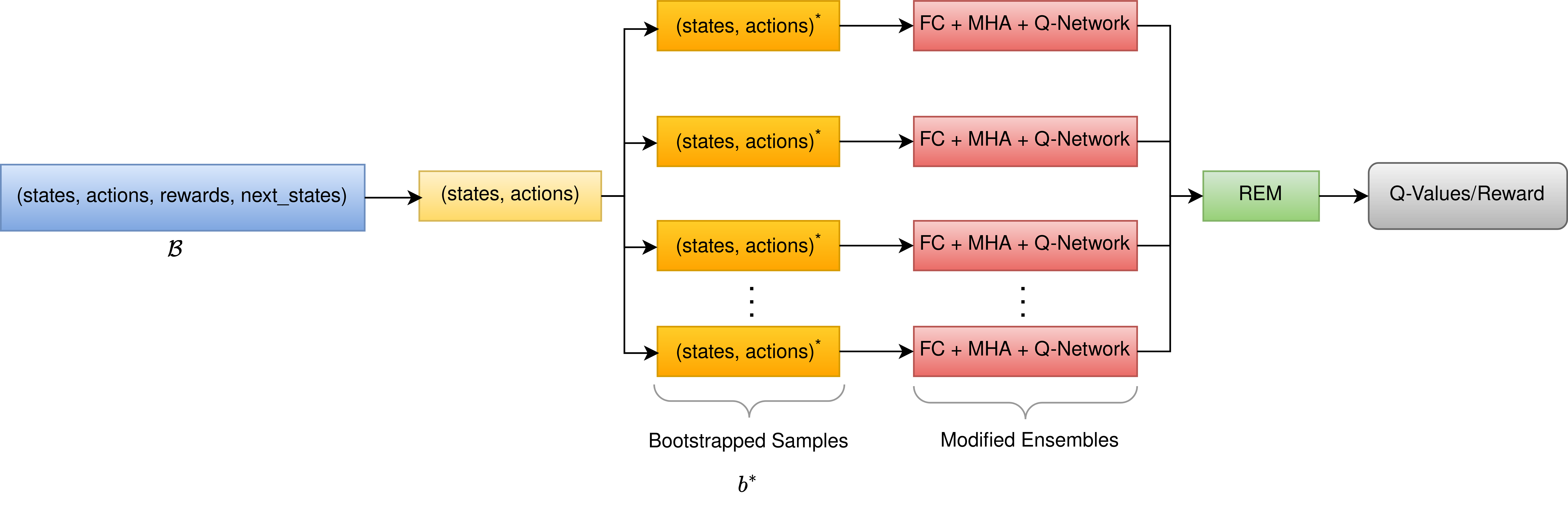}
    \caption{Our modified REDQ approach incorporates both the bootstrapping and MHA mechanisms. Both the states and the actions are first concatenated, and then multiple bootstrapped samples are drawn from the replay buffer for the ensemble of Q-learners.  Individual Q-learners incorporate a fully connected layer and multi-head attention on top of a Q-network.  The Q-network integrates elements from both the REDQ and DroQ implementations.}
    \label{fig:fig1}
\end{figure*}

\section{Related Work}

Model-free deep RL approaches such as TRPO \cite{trpo}, PPO \cite{ppo} and A3C \cite{a3c} have been applied to a variety of decision making and control tasks. While these approaches provide reasonable performance, they suffer from poor sample efficiency due to on-policy learning, requiring new sampling at each step. To tackle these challenges, Soft Actor Critic \cite{sac}, an off-policy learning method, was proposed.  SAC achieves a higher sample efficiency, but still uses a very low update-to-data (UTD) ratio which somewhat limits the potential sample efficiency.

In contrast, MBPO \cite{mbpo} represents a model-based approach that manages the trade-offs between leveraging a model to generate data and the risks of using an inaccurate model. It integrates both real data from the environment and synthetic data generated from the model, and utilizes higher UTD ratio, usually 20 to 40, to achieve better sample efficiency. 

Given the success of MBPO with a high UTD ratio, many recent deep RL methods use a high UTD ratio to achieve better sample complexity than model-based deep RL methods \cite{mbpo_unsup,bmbpo}. However, the higher UTD ratio comes at the cost of overestimation bias.

In the ongoing quest to improve the performance of deep RL methods, researchers have proposed diverse approaches. Notable strategies include ensembles of Q-functions \cite{ensemble1,ensemble2,ensemble3}, integration of dropout transition models \cite{drop1,drop2}, and application of normalization techniques such as batch normalization \cite{bnorm} and layer normalization \cite{layernorm}. While previous methods employed ensembles to capture model uncertainty in both target calculation and policy optimization, they did not specifically focus on the issue of overestimation.

To address the overestimation problem caused by the higher UTD ratio, \citeauthor{redq} (\citeyear{redq}) proposed Randomized Ensembled Double Q-learning (REDQ).  This model-free approach attains superior performance and similar sample efficiency to MBPO, mainly by using a higher UTD ratio. It mitigates the overestimation bias resulting from the increased UTD ratio by employing a large ensemble of Q-functions and then choosing a random subset of the Q-function ensembles for in-target minimization. 

\citeauthor{droq} (\citeyear{droq})
were able to improve the computational performance of REDQ.  To achieve this improvement, DroQ incorporates the techniques such as dropout \cite{dropout} and layer normalization \cite{layernorm}. Moreover, it aims to discover a policy capable of maximizing the expected return while incorporating an entropy bonus.
    

\section{Method}
Our method incorporates multi-head self-attention, identity connections, and bootstrapping into the REDQ architecture.
Our modifications are shown in Figure \ref{fig:fig1} while the pseudocode is presented in the Algorithm \ref{alg1}.  Differences between the algorithms are highlighted in red.  
\begin{algorithm}[H]
\centering
    \caption{Modified REDQ}\label{alg1}
    \begin{algorithmic}[1]
        \State Initialize policy parameters $\theta$, $N$ Q-function parameters $\phi_{i}, i = 1,...,N$, empty replay buffer $\mathcal{D}$. \\
        Set target parameters $\phi_{tar,i} \leftarrow \phi_i$, for {$i = 1, 2, ..., N$} 
        \Repeat
            \State Take an action $a_{t} \sim \pi_{\theta}(.|s_{t})$ and observe reward $r_{t}$, next state $s_{t+1}$
            \State Update buffer $\mathcal{D} \leftarrow \mathcal{D} \bigcap (s_{t}, a_{t}, r_{t}, s_{t+1})$
            \For{$\mathbf{G}$ updates}
            \State Sample a mini-batch $\mathcal{B} = {(s, a, r, s^{'})}$ from $\mathcal{D}$
            \color{red} 
            \For{$b_i$ in $\mathcal{B}$}
            \State Select a sub-sample $b^*$ from mini-batch $\mathcal{B}$ with replacement
            \State Apply multi-head self-attention to bootstrapped sample $b^*$
            \[
                \text{Attention} (q, k, v) = \text{softmax} \left( \frac{qk^{T}}{\sqrt{d_{k}}} \right) v
            \]
            \State Calculate Q for all $\mathcal{M}$ of $\mathbf{M}$ distinct indices from ${1, 2, ..., N}$ with bootstrapped sample $b^*$
            \color{black}
            \State Calculate target $y$ as:
            
            \begin{align*}
                y = r + \gamma \left( \min_{i \in \mathcal{M}} Q_{\phi_{tar, i}}(s^{'}, \tilde{a}{'}) - \alpha \log\pi_{\theta}(\tilde{a}' | s^{'}) \right),  \\
                \tilde{a}' \sim \pi_{\theta}(.|s^{'})
            \end{align*}
            \color{red}
            \EndFor
            \color{black} 
            \For{$i = 1, ..., N$} 
            \State Update $\phi_{i}$ with gradient descent using
            \[
                \bigtriangledown_{\phi} \frac{1}{|\mathcal{B}|} \sum_{(s, a, r, s^{'})} (Q_{\phi_{i}}(s, a) - y)^{2}
            \]
            \State Update target networks with 
            \[
                \phi_{tar, i} \leftarrow \rho \phi_{tar, i} + (1 - \rho) \phi_{i}
            \]
            \EndFor
            \EndFor
            \State Update policy parameters $\theta$ with gradient ascent using
            \begin{gather*}
                 \bigtriangledown_{\theta} \frac{1}{|\mathcal{B}|} \sum_{(s \in \mathcal{B})} \left( \frac{1}{N} \sum_{i=1}^{N} Q_{\phi_{i}}(s, \tilde{a}_{\theta}(s)) - \alpha \log \pi_{\theta}(\tilde{a}_{\theta}(s) | s) \right), \\
                 \tilde{a}_{\theta}(s) \sim \pi_{\theta}(.|s)
            \end{gather*}
         \Until{done}
    \end{algorithmic}
\end{algorithm}
Similar to both REDQ and DroQ, our approach involves utilizing $N$ ensembles of Q-functions, and we select a subset $\mathcal{M}$ out of $N$ for in-target minimization. Each Q-network is constructed using the original implementation of the REDQ method. We also integrate the key modifications inspired by the DroQ, incorporating a dropout layer and layer normalization into each Q-network. 

Moving forward, as depicted in Figure \ref{fig:fig1}, we introduce our modifications to each Q-network described above. Each Q-network now boasts a fully connected layer and a multi-head self-attention layer, carefully positioned before the original Q-network implementation. This augmentation ensures the model is capable of capturing even richer dependencies and improves the overall expressive power of the model. 
During the training process, at each step, we draw a mini-batch $\mathcal{B}$ from the replay buffer $\mathcal{D}$ and generate multiple bootstrapped samples $b^{*}$ from $\mathcal{B}$. We evaluated different sizes of the bootstrap sample, $|b^*|=\{2, 4, 8\}$; results are reported for $|b^*|=4$. Each sample is comprised of concatenated state-action pairs.  The incorporation of bootstrapped samples serves several crucial purposes: 1) these samples help agent learn from both its own prediction and those generated by the Q-network; 2) the agent updates its policy based on information it has while reducing the need for extensive exploration to gather new experiences and hence reducing environment interactions; 3) as the agent potentially encounters some state-action pairs multiple times, this repetition significantly amplifies the agent's predictive capacity; 4) bootstrapping also reduces the variance. 

Each bootstrapped sample $b^{*}$ then undergoes transformation through a fully connected layer before being fed to the multi-head self-attention layer. This transformation ensures that all samples are in the same representation space and share a consistent embedding dimension, necessary to apply self-attention. The multi-head self-attention is calculated as given by Equations \ref{eq:head} and \ref{eq:attn}.
\begin{gather} 
    \label{eq:head}
     head_{i} = \text{Attention}(q, k, v) = \text{softmax} \left( \frac{qk^{T}}{\sqrt{d_{k}}} \right) v 
    \\
    \label{eq:attn}
    \text{MultiHead}(q, k, v) = \text{concatenate}(head_{i}, ..., head_{H})w
\end{gather} 
where the matrices \textbf{k}, \textbf{v}, and \textbf{q} $\in \mathbb{R}^{\mathcal{N} \times d}$ are calculated from each $b^{*}$ sub-sample. The $d_{k}$ is the embedded dimension of the model set to either $d \in \mathbb{R}^{256}$ or $d \in \mathbb{R}^{512}$, depending upon the environment. $H$ is the number of heads which is set to $8$ for most of the environments.  

\begin{figure*}[t!]
    \centering
    \includegraphics[width=1\linewidth]{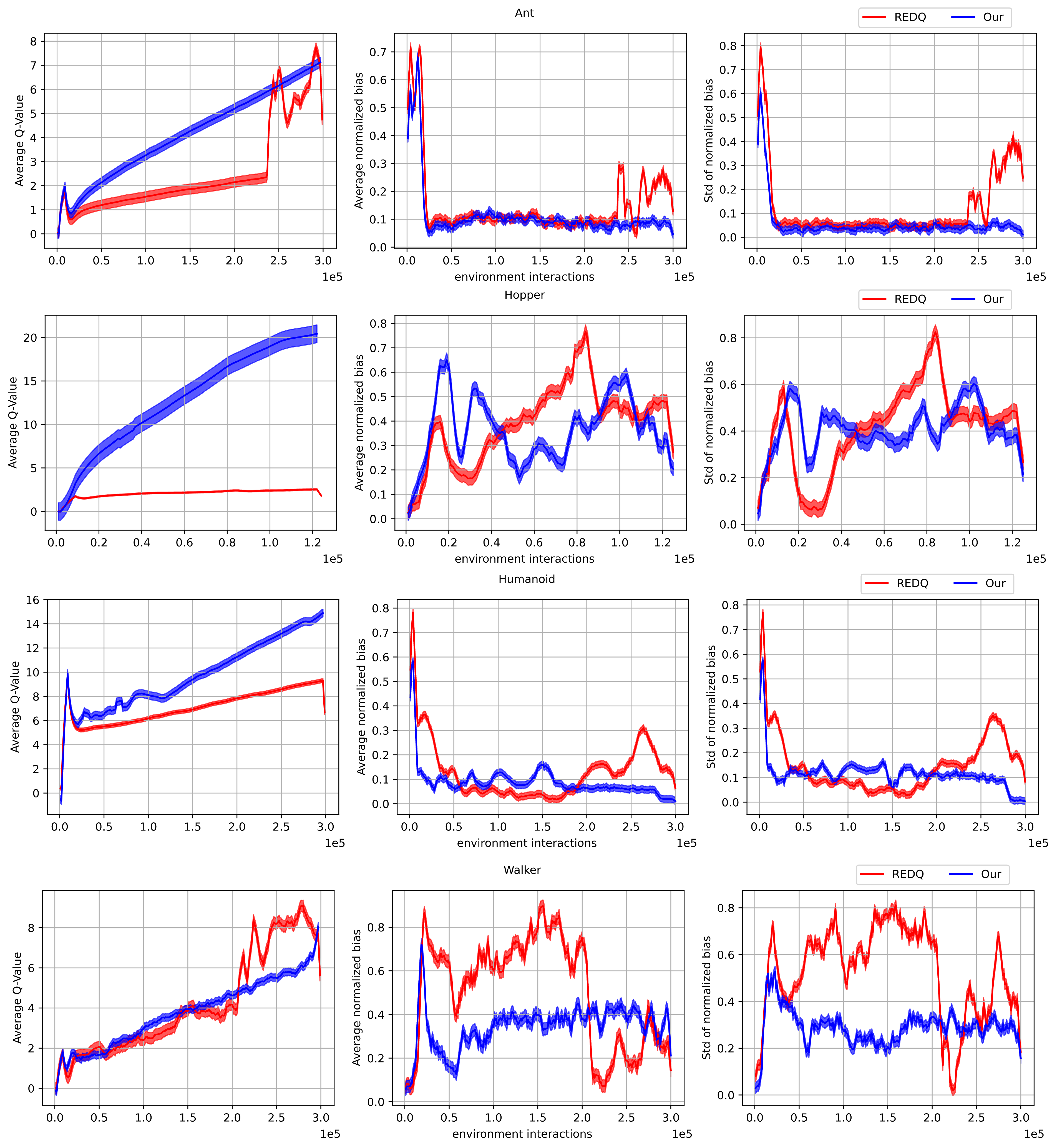}
    \caption{An evaluation of Q-value prediction, average normalized Q-bias, and standard deviation of normalized Q-bias: REDQ vs. our proposed approach. Results are based on three separate runs for each environment. Our method demonstrates enhanced Q-value predictions while effectively managing estimation bias at a level comparable to REDQ.}
    \label{fig:2}
\end{figure*}


The MHA provides significant advantages: 1) by applying attention to state-action pairs, the agent gains a better understanding of the relationship between states and actions, thereby enhancing temporal modeling; 2) it also aids the agent in optimizing its policy and making informed decisions while taking
actions; 3) leveraging MHA, each head has the capacity to learn a distinct and meaningful relationships for a state-action pair; 4) and lastly, the MHA ensures that the state-action pairs are invariant to permutations.  

While the combination of MHA and bootstrapping contributes to performance improvements over REDQ, in the case of DroQ, MHA typically does not enhance performance. Instead, it often leads to a declined performance. This discrepancy can be attributed to the simplicity of the DroQ network, where the inclusion of attention mechanisms adds unnecessary complexity, adversely affecting performance.

To address this issue, we propose a solution by incorporating identity connections, inspired by the concept introduced in ResNet \cite{resnet}. By removing MHA and integrating identity connections while retaining the bootstrapping approach, we effectively streamline the network architecture and mitigate the performance issues associated with unnecessary complexity.


In the REDQ framework, the ensemble size is typically set to $N = 10$, with a fixed random subset size of $M = 2$ and UTD is set to $20$. To ensure a fair comparison in our initial experiments, we adopt the same configuration as outlined by the REDQ implementation, aligning our setup with their established parameters.

\begin{figure*}[t!]
    \centering
    \includegraphics[width=1\linewidth]{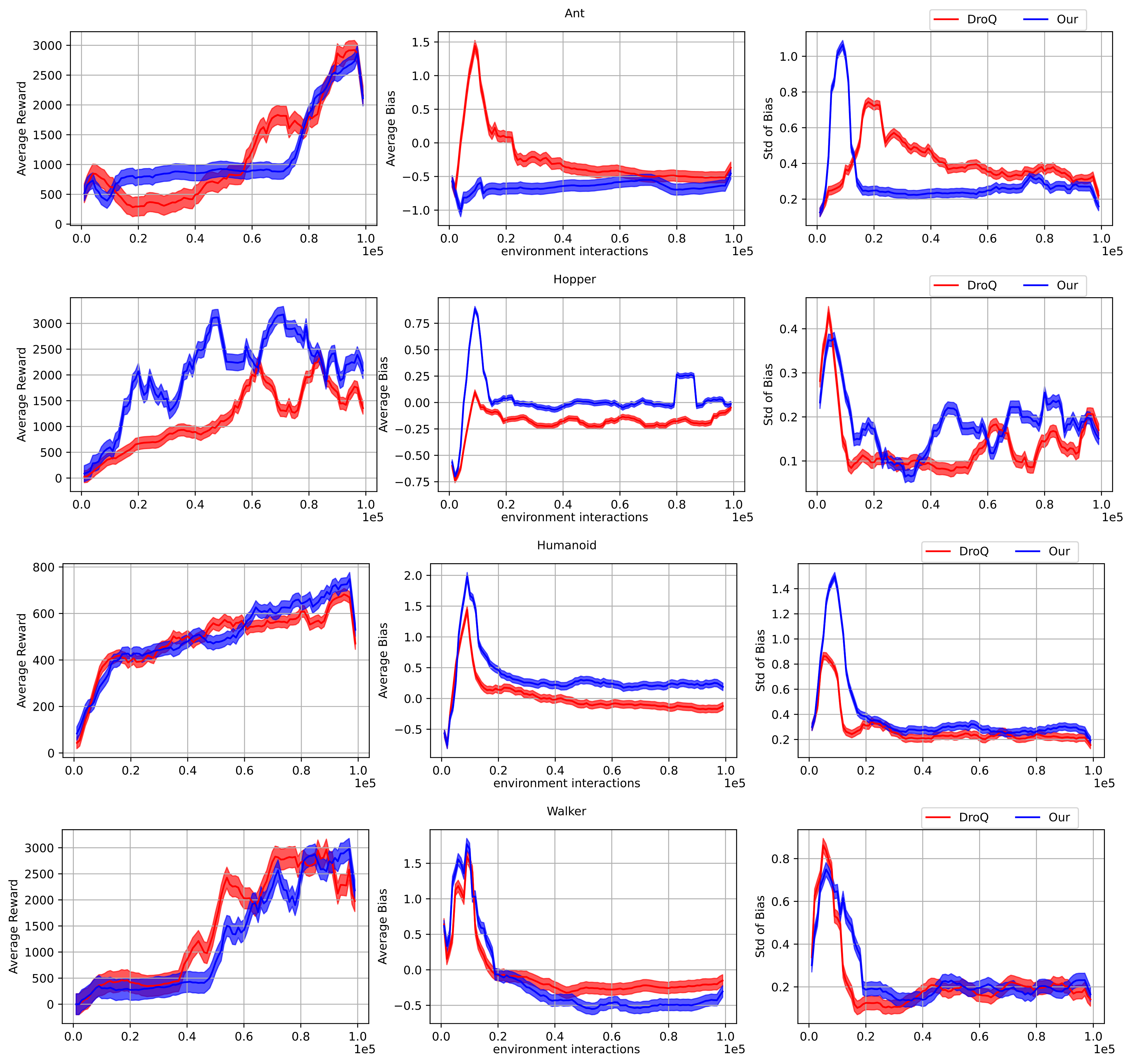}
    \caption{Comparison of our method with identity connections and bootstrapping vs.\ DroQ. Our approach improves over DroQ in all the environments. Results were obtained from three distinct runs.}
    \label{fig:skip}
\end{figure*}

However, for our subsequent experiments, we diverge from the conventional REDQ configuration. We choose an ensemble size of $N = 5$, while still maintaining the subset size at $M = 2$ and reducing the UTD from $20$ to $10$. In both configurations, each individual Q-network is initialized randomly, but they all undergo updates with the same target. In each case, the target value is determined as follows:

\begin{gather*}
    \mathbb{E}_{\bar{\phi}_{i}, .., \bar{\phi}_{M}} \left[ \min_{i = 1, .., M} Q_{\phi_{tar,N}}(s', a') \right] \approx \min_{i=1, .., N} Q_{\phi{tar, i}}(s', a')
\end{gather*}

\begin{figure*}[!htp]
    \centering
    \includegraphics[width=.95\textwidth]{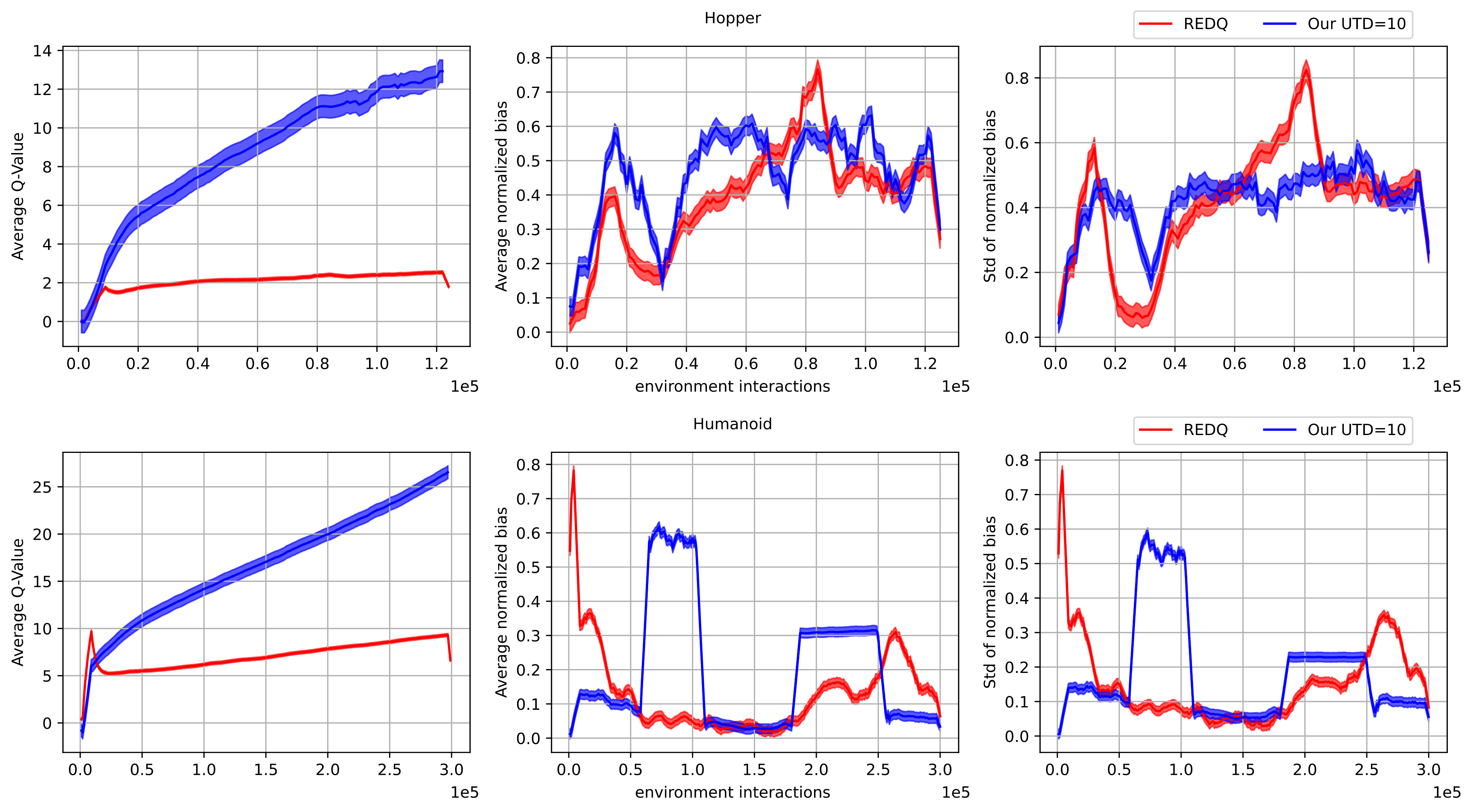}
    \caption{Comparison of Q-value prediction, average normalized bias, and standard deviation of normalized bias between REDQ ($G = 20, N = 10$) and our approach when UTD is set to $G = 10$ while $N = 5$ for our approach. Our method still managed to achieve better performance than REDQ.  Results were obtained from three distinct runs for each environment.}
    \label{fig:low_utd}
\end{figure*}

where each Q-function of $N$ is independently initialized with parameters $\bar{\phi}_{i}, .., \bar{\phi}_{N}$ which is given by line 12 of the algorithm. Our approach balances the dual aims of providing the model with ample uncertainty without being excessively computation intensive. It delivers consistent performance, even when the UTD is reduced $( UTD < 20)$ (Figure \ref{fig:low_utd}).

\subsection{Experimental Results}
We showcase our findings across four challenging OpenAI Gym environments: Ant-v2, Hopper-v2, Humanoid-v2, and Walker2d-v2.  In addition to analyzing the rewards, we also look at average estimation bias and the standard deviation of estimated bias. The estimation bias is calculated as:
\[
    Q_{\phi}(s, a) - R^{\pi}(s, a)
\]
where $R^{\pi}(s, a)$ is the Monte Carlo return while $Q_{\phi}(s, a)$ is the average return of randomly selected subset of Q-learners. To better predict the estimation bias, normalized estimation bias is also used as a metric and is calculated as: 
\[
    b(s, a) = (Q_{\phi}(s, a) - R^{\pi}(s, a)) / \mathbb{E}_{s, a\sim \pi}[R^{\pi}(s, a)]
\]

Both the mean and standard deviation of estimation bias are crucial metrics in assessing the performance. A high absolute mean bias indicates that the estimates are consistently inaccurate. On the other hand, a high standard deviation of bias suggests that the estimation errors are not uniform across different states or actions.  Our aim is to have the average normalized bias fall close to zero with a small standard deviation.

These experiments are conducted using the optimal parameters of the REDQ. In this comparison, following the REDQ approach, we choose $N = 10$ and $M = 2$, while the UTD is set to $G = 20$. The results of this comparison are illustrated in Figure \ref{fig:2}.  Under this configuration, our method achieves superior average Q-value predictions while simultaneously keeping estimation bias at a minimum. Furthermore, our approach outperforms REDQ by maintaining better average normalized bias and standard deviation of normalized bias. These results affirm that our method is not only proficient in handling overestimation and underestimation but is, in fact, at least as effective as REDQ. 


We conducted another set of experiments using the optimal configuration for both DroQ and our method, incorporating identity connections, with parameters set to $N = M = 2$ and $G = 20$. The results are illustrated in Figure \ref{fig:skip}. Our method, enhanced by bootstrapping and identity connection, consistently achieves superior rewards across all environments, effectively mitigating estimation bias by keeping both average bias and standard deviation of average bias close to zero. 

A higher UTD ratio has substantial challenges including overfitting to specific experiences, constraint exploration, reducing the generalization and increasing the computational cost. One of the straightforward ways to address these challenges is to reduce the UTD ratio. In our subsequent set of comparisons, we opt for a more aggressive configuration by choosing $N = 5$ and $M = 2$ while reducing the UTD $G = 20$ to $G = 10$. For REDQ, we leave the values of $N$ and $M$ unchanged. These results are presented in Figure \ref{fig:low_utd}.
In this set of experiments, the results exhibit a somewhat mixed nature. Although the average Q-values prediction is reduced for both methods, our approach still manages to perform equally well in terms of average normalized Q bias and standard deviation of normalized Q bias. 

%

\section{Conclusion}

This paper introduces an enhanced variant of the REDQ method that incorporates bootstrapping and multi-head self-attention into an ensemble of Q-functions.  Our method demonstrates enhanced Q-value predictions while effectively managing estimation bias at a level comparable to REDQ, even when reducing the UTD ratio and ensemble size.  A version of our model outperforms the optimal configuration for DroQ. Our results demonstrate the benefits of our proposed approach towards improving both sample efficiency and performance across multiple OpenAI Gym environments.  



\clearpage
\bibliographystyle{flairs}
\bibliography{ref}
\clearpage
\onecolumn
\section*{Appendix}

This section provides additional results from our method in comparison to REDQ.



\begin{figure}[!ht]
    \centering
        \includegraphics[width=0.8\linewidth]{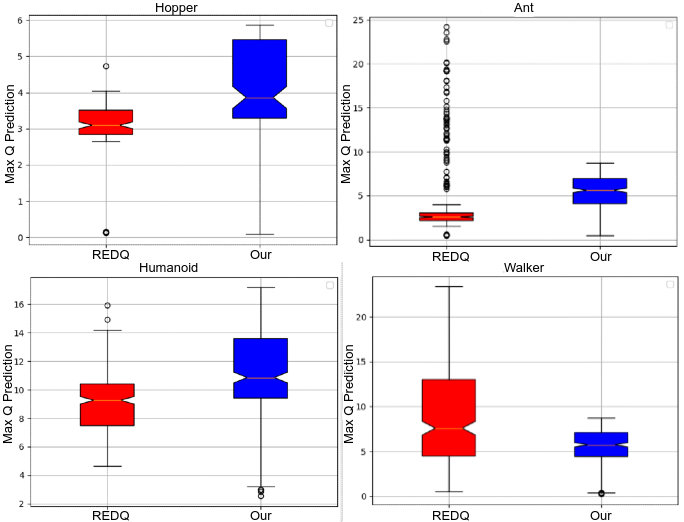}
        \caption{Comparison of max Q-values prediction distribution. Our method achieves better prediction distribution while simultaneously reducing the number of outliers, except in the case of Walker.}
        \label{fig:maxq}
\end{figure}

Figure~\ref{fig:maxq} shows the distribution of maximum Q values for both methods.  With the same optimal configuration, our method also achieves better maximum Q-values distribution for all environments, except Walker2d. Our approach exhibits fewer outliers compared to REDQ. For instance, in the case of Ant-v2, the REDQ generates an excessive number of outliers, whereas our approach significantly reduces these outliers. 
\newpage
\begin{figure}[!ht]
\centering 
        \includegraphics[width=0.8\linewidth]{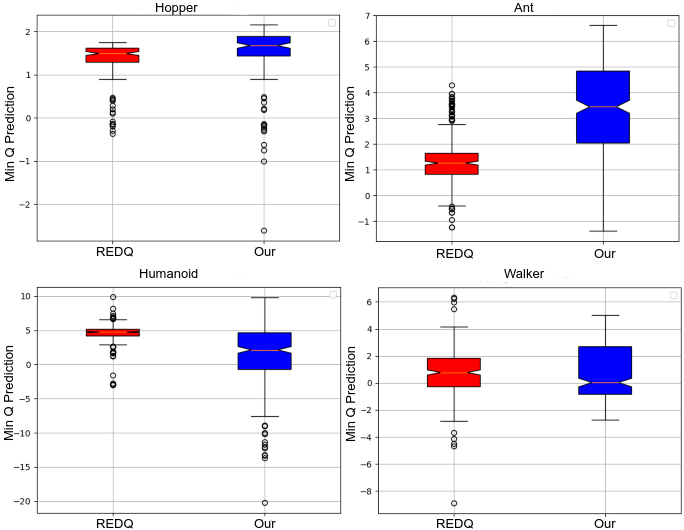}
        \caption{Comparison of min Q-values prediction distribution. Once again, the distribution demonstrates that our method is better than REDQ at Q-value prediction, effectively reducing the number of outliers.}
        \label{fig:minq}
\end{figure}

Figure~\ref{fig:minq} shows the minimum Q-values distribution for both methods. Our proposed technique not only outperforms the REDQ in reducing outliers across multiple environments but also exhibits good Q-value prediction, except in the Humanoid-v2 environment. Although the minimum Q-values prediction of REDQ is better on Humanoid-v2, our method still reduces the number of outliers.
By achieving overall improved distributions for both maximum and minimum Q-values prediction across the environments, we conclude that our approach is effective at controlling for overestimation and underestimation bias. 
\newpage
\textbf{Effects of number of $\mathbf{b^{*}}$}

\begin{figure}
    \centering
    \includegraphics[width=0.5\linewidth]{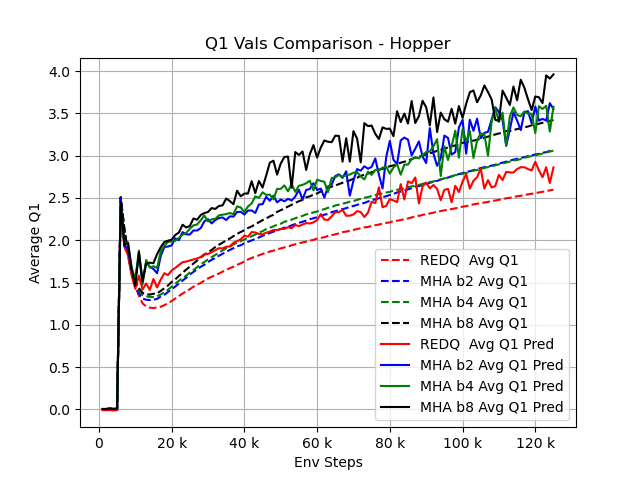}
    \caption{Effects of number of $b^*$ samples}
    \label{fig:effectsb}
\end{figure}

We conducted experiments using different numbers of $b^*$ samples i.e., \(b_i = \{2, 4, 8\} \). Empirically, we found that \( b_i = 4 \) yielded superior results, as depicted in Figure \ref{fig:effectsb}. Although \( b_i = 8 \) produced higher Q1 values, it also introduced increased estimation bias compared to \( b_i = 4 \). In all these experiments, we kept the batch size to $\mathcal{B} = 512$.

\newpage
\textbf{Effects of Hidden Dimension}

\begin{figure}
    \centering
    \includegraphics[width=0.5\linewidth]{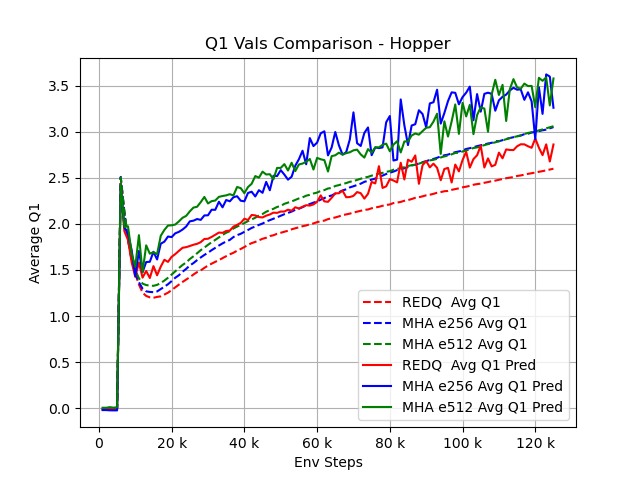}
    \caption{Effects of Hidden Dimension}
    \label{fig:hiddendim}
\end{figure}

We also experimented with various hidden dimensions for both the Q-networks and multi-head self-attention while keeping the $b^*$ samples to $b_i = 4$. We found that the hidden dimension $d \in \mathbb{R}^{512}$ performed better and achieved higher values while keeping the estimation low as depicted in figure \ref{fig:hiddendim}.

\end{document}